\documentclass[journal]{IEEEtran}

\usepackage[noadjust]{cite}

\usepackage{caption}
\usepackage[export]{adjustbox}
\usepackage{graphicx}
\usepackage[caption=false]{subfig}
\graphicspath{{figures/}}

\usepackage{siunitx}
\usepackage{etoolbox}

\usepackage{float}
\usepackage{comment}
\usepackage{csquotes}

\ifCLASSINFOpdf

\else

\fi

\usepackage{amsmath,amsfonts,amssymb}
\usepackage[table,xcdraw]{xcolor}
\usepackage{multirow}
\makeatletter
\newcommand*{\rom}[1]{\expandafter\@slowromancap\romannumeral #1@}
\makeatother

\begin{document}

\title{Vehicle Teleoperation: Successive Reference-Pose Tracking to Improve Path Tracking and to Reduce Time-Delay Induced Instability}

\author{Jai~Prakash,~Michele~Vignati,~Edoardo~Sabbioni,~and~Federico~Cheli

\thanks{The authors belong to the Department of Mechanical Engineering, Politecnico Di Milano, Italy (e-mail:
jai.prakash@polimi.it; michele.vignati@polimi.it; edoardo.sabbioni@polimi.it; federico.cheli@polimi.it)
}%

}


\maketitle


\begin{abstract}

Vehicle teleoperation is an interesting feature in many fields. A typical problem of teleoperation is communication time delay which, together with actuator saturation and environmental disturbance, can cause a vehicle deviation from the target trajectory imposed by the human operator who imposes to the vehicle a steering wheel angle reference and a speed/acceleration reference.


With predictive techniques, time-delay can be accounted at sufficient extent. But, in presence of disturbances, due to the absence of instantaneous haptic and visual feedback, human-operator steering command transmitted to the the vehicle is unaccounted with disturbances observed by the vehicle. To improve reference tracking without losing promptness in driving control, reference trajectory in the form of successive reference poses can be transmitted instead of steering commands to the vehicle. 

We introduce this new concept, namely, the \enquote{successive reference-pose tracking (SRPT)} to improve path tracking in vehicle teleoperation. This paper discusses feasibility and advantages of this new method, compare to the smith predictor control approach. 

Simulations are performed in SIMULINK environment, where a 14-dof vehicle model is being controlled with Smith and SRPT controllers in presence of variable network delay. Scenarios for performance comparison are low adhesion ground, strong lateral wind and steer-rate demanding maneuvers. Simulation result shows significant improvement in reference tracking with SRPT approach.

\end{abstract}

\begin{IEEEkeywords}
Latency, time-delay, vehicle teleoperation, smith predictor, NMPC, successive reference-pose tracking.
\end{IEEEkeywords}

\section{Introduction}
%
%
%
%

\IEEEPARstart{V}{ehicle teleoperation} refers to driving a vehicle by transmitting driving commands to the vehicle's control system from a control station that is stationary and, in general, far away from the vehicle. A human operator who remains in the control station generates the driving commands. The data communication protocol, interestingly, can be wired or wireless. Wireless data communication protocol is required due to the potential applications of vehicle teleoperation, which include last-mile delivery of rental/shared vehicles, avoiding driver presence in dangerous areas, human remote assistance in the event of autonomous vehicle failure, and valet parking, among others. Due to its widespread availability around the world, 4G LTE wireless broadband connection is the greatest candidate for communication. Fully autonomous vehicles may perform well in the aforementioned scenarios, but they may still fail in key traffic situations that a human could easily handle, such as parking lots, pedestrian crossing areas, or construction roads. Vehicle teleoperation could potentially help with the transition from human-driven to completely autonomous vehicles.

Vehicle teleoperation has a number of challenges, including control loop delays, complete connection loss, limited situation awareness, etc. Human performance in virtual environments has shown that people can notice latency as low as 10–20 ms. \cite{Ellis2004}. According to control theory, time delays can cause the vehicle's real-time control to become unstable \cite{Pongrac2011GestaltungUE}. Frank L. H. \cite{Frank1988}, in a simulated driving task, observed that delays of 170 milliseconds had a significant impact on driving performance. Humans can adapt to time delays in control systems, but this adaptation is contingent on the human operator's capacity to foresee the outcome of his control actions, and the level of this adaptation is determined by the time delay's features (e.g., magnitude and variability)\cite{Davis2010}. Human operators, on the other hand, have a tendency to overcorrect steer when delays are large, resulting in oscillations that can significantly reduce teleoperation performance and even destabilize the closed-loop system \cite{Sheridan1993}.

Literature of interest can be classified broadly in two major categories: Predictive display and Trajectory based vehicle teleoperation. Predictive displays, have been proven effective to compensate delays and improve vehicle mobility in human-in-the-loop experiments \cite{Chucholowski, 7504430, Brudnak2016PredictiveDF, jai2022, tito2015, Zheng2019, Zheng2020}. It considers the delay together with the control signals from the operator to estimate the vehicle position. This predicted position is displayed to the operator as a "third-person view" \cite{tito2015}. The prediction model can either be model-based \cite{jai2022}, model-free \cite{Zheng2019}, or a blend of both \cite{Zheng2020}. In model-based predictors, a vehicle model is required to predict the vehicle response, and the prediction accuracy depends on the accuracy of the vehicle model. In the presence of unknown disturbances in driving scenario such as low adhesion road or cross-wind, prediction accuracy deteriorates. In model-free approach \cite{Zheng2019}, prediction is made by taking into account the delayed state dynamics received from the vehicle. It suffers from convergence time in the state prediction which results in delayed prediction. Blending of both results improved operation, but not much significantly. In summary, predictive displays tries to bypass the time delay in loop, by predicting states taking input the delayed states. This is helpful for human-in-the-loop teleoperation, as it allows the human operator to not to wait for the feedback and provides the sense of controlling the vehicle in real-time. The limitation is, if the prediction accuracy decreases, chances of asynchrony increases.

In trajectory-based driving (shared-control approach), vehicle control is based on automated driving along predefined paths, using trajectories as command variables, which consist of parameterized curves overlaid by reference vehicle speeds. This eliminates network time-delay out of control loop. Control-station transmits the mission plan, and vehicle autonomy performs the maneuvers. Michael Fennel\cite{Fennel2021}, proposed an offline path follower where the operator “draws” a desired 2D path by walking in a large-scale haptic interface while a guiding force is exerted, which ensures that the generated path can later be accurately followed by a path tracking controller running offline on a remote robot. The limitation of trajectory-based driving is, the operator is not actively controlling the task.

\begin{figure}[h]
    \centering
    \includegraphics[width=0.37\textwidth]{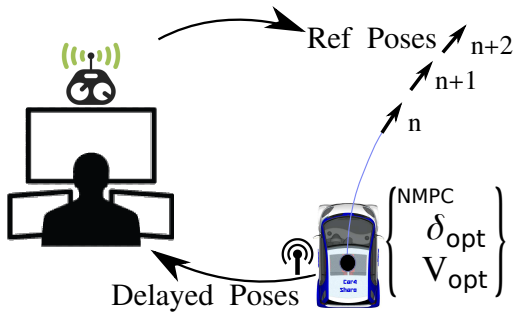}
    \caption{Successive reference-pose tracking (SRPT) in vehicle teleoperation}
    \label{fig:x vehicleTeleoperation}
\end{figure}

This paper extends the trajectory-based driving approach, according to fig (\ref{fig:x vehicleTeleoperation}), by transmitting successive reference-pose to the vehicle at $30$ fps instead of transmitting the whole mission plan. The successive poses corresponds to the target to reached in around $1$s. A Non-linear Model Predictive Control (NMPC) inside the vehicle optimizes steer and vehicle speed considering vehicle states, steer actuator constraints, and vehicle acceleration limits adopting a conservative friction value according to the reference pose sent by the control-station. The contribution of this paper is the introduction of the approach of Successive Reference-Pose Tracking (SRPT), in which control-station transmits target 2D-poses instead of direct steer and speed commands. To assess its usefulness, result of this approach is compared with state prediction (Smith predictor) approach in Simulink simulation environment. In simulation environment, variable network delays and 14-dof vehicle model for the main vehicle are considered.

The paper is organized as follows. Section II - Method, A) exhibits delay trend in vehicle teleoperation over 4G. B) discuss the Smith predictor approach, C) explains the SRPT approach. Section III compares the two approaches over their simulation results. Section IV concludes the paper with key findings.
\section{Method}
\subsection{Uplink delay and variable downlink delay}
\begin{figure}[h]
\centering
    \includegraphics[width=0.9\columnwidth]{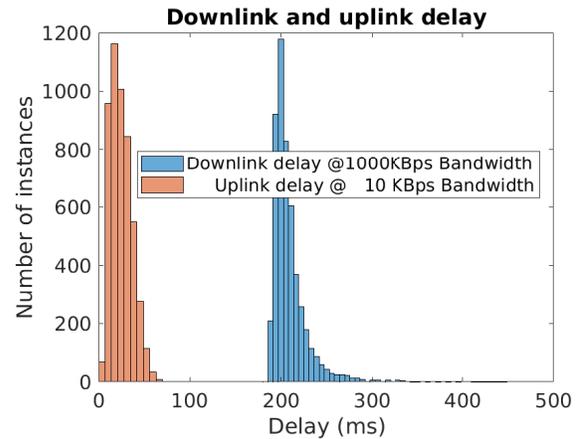}
    \caption{Delays observed in data transmission over 4G}
    \label{fig: delays}%
\end{figure}

Taking control-station as reference point, time-delay in vehicle teleoperation can be divided into two parts. One is downlink-delay ($\tau_2$), associated to the availability of streamed images to the human operator at the control station. Other is uplink-delay ($\tau_1$), associated to the delay between generation of driving commands at the control station and actuation of those at the vehicle. $\tau_2$ can be considered as the sum of camera exposure delay, image encoding time consumption, network delay in transmitting the images towards the control station and image decoding time consumption. These contributions can be lumped, because time point of image capture is used in image time-stamping. $\tau_1$ can be considered as the sum of network delay in transmitting the driving commands towards the vehicle, and vehicle actuation delay. In case of wireless communication using 4G, variability is associated both downlink and uplink delays. Fig(\ref{fig: delays}) presents both sided delays with corresponding utilized bandwidth. Here the data corresponds to 5000 image frames and corresponding driving commands. This test is performed in typical urban environment where the vehicle is connected to 4G mobile communication and control-station is connected to internet using wired LAN.

From the control-station perspective, uplink-delay is unknown and the downlink-delay is known. Being smaller in amplitude and relatively lower variability of uplink-delays, a constant high stochastic value (an approach adopted also in \cite{jai2022}) of 60ms is considered in teleoperation simulations of this paper. The downlink delay is known, as the data packets are timestamped. \emph{Generalized extreme value distribution}, $GEV(\xi=0.29, \mu=200, \sigma=9)$ fits accurately on downlink-delay experimental data. Also, authors from articles \cite{Zheng2020, jai2022} found GEV distribution apt for representing distribution of time-delays in  mobile communication. Here, $\xi$ is shape parameter, $\mu$ is location parameter and $\sigma>0$ is scale parameter. Positive $\xi$ means, the distribution has a lower bound $(\mu-\frac{\sigma}{\xi})\approx169ms\:(>0)$ and a continuous right tail, based on the extreme value theory. During a stable 4G connection (fig \ref{fig: delays}), data packets received are found to be in FIFO order. Which indicates no data loss and FIFO queue behaviour of the communication. 

\subsection{Smith predictor in vehicle teleoperation}

\begin{figure}[h]
\centering
    \includegraphics[width=\columnwidth]{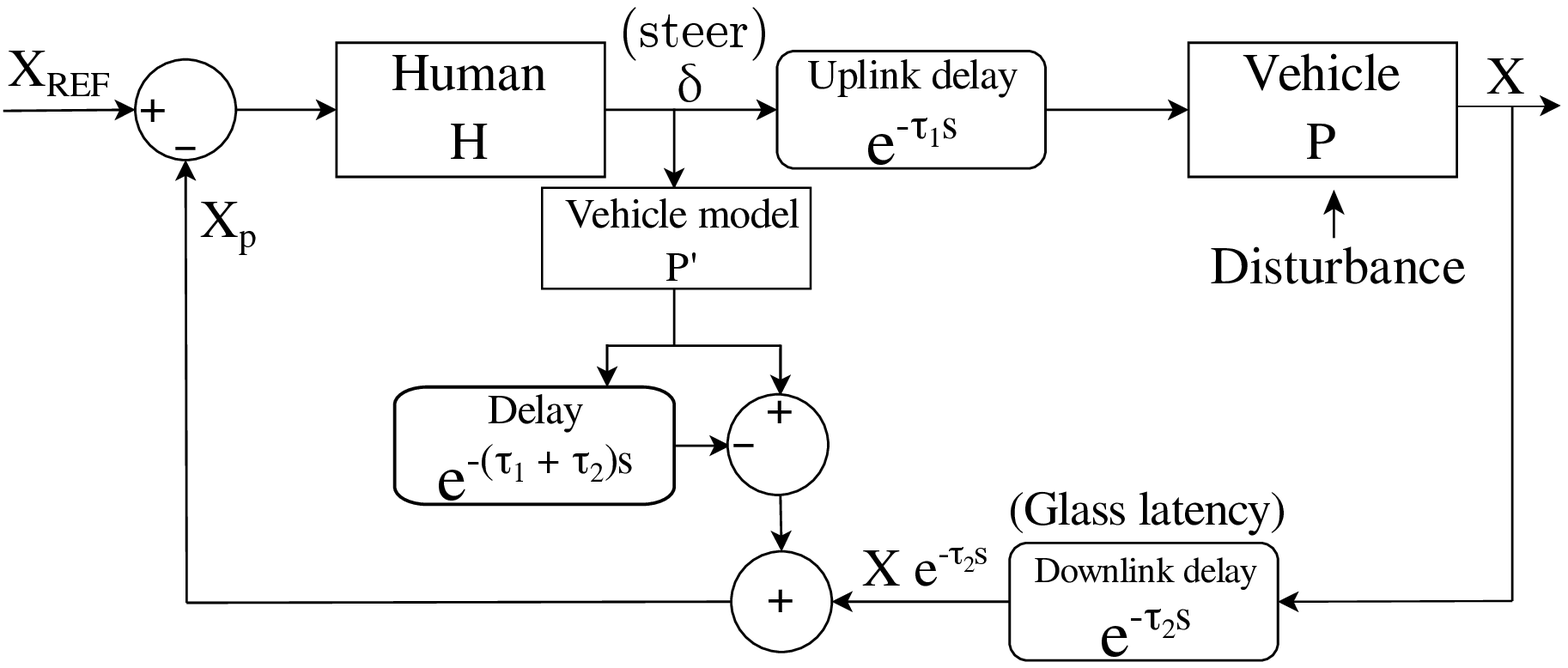}
    \caption{Smith predictor scheme}
    \label{fig:smith_1}%
\end{figure}

Among the predictive control approaches in bilateral teleoperation, Smith predictor approach is quite popular. It is a model based prediction approach introduced by O.J. Smith in 1957 \cite{smith}. A schematic of the Smith Predictor in vehicle teleoperation control loop is shown in fig (\ref{fig:smith_1}) for system with known time-delays. The control input ($\delta$) is passed through a local predictor model ($P^{\prime}$) of the vehicle, which then passes through $(1-e^{-\left(\tau_{1}+\tau_{2}\right) s})$, where a time-delayed version of the output is subtracted from the real-time version. With this schematic, feedback given to the human operator ($X_p$) is 

\begin{equation}
X_p = P^{\prime}u\left(1-e^{-\left(\tau_{1}+\tau_{2}\right) s}\right)+\delta e^{-\tau_{1} s} P e^{-\tau_{2} s}\label{eq smith3Feedback}
\end{equation}

Which in turn if the predictor model ($P^{\prime}$) is equal to vehicle model ($P$) becomes
\begin{equation}
X_p = P\delta\label{eq smith4Feedback}
\end{equation}

It provides the human operator the sense of controlling the vehicle in real-time. Although the Smith Predictor was originally aimed at controller design, the same scheme can be used to help the human better observe the feedback state by eliminating the backward delay. The transfer function of the closed loop delayed system is 

\begin{equation}
\frac{X}{X_{REF}}=\frac{H P e^{-\tau_{1}s}}{1+H P^{\prime}-H P^{\prime} e^{-\left(\tau_{1}+\tau_{2}\right)s}+H P e^{-\left(\tau_{1}+\tau_{2}\right)s}}\label{eq smith1}
\end{equation}
\\
If $P^{\prime}=P$, it becomes

\begin{equation}
\frac{X}{X_{REF}}=\frac{H P }{1+H P}e^{-\tau_{1}s}\label{eq smith2}
\end{equation}

Smith Predictor results bypassing the delay in the observation and transform the system into a pure forward delay to the vehicle. The backward delay is eliminated, and the forward delay is moved aft of the control loop. This allows the human operator to command the vehicle without being hindered by the time delay. The system still tracks the input with a constant forward delay offset; however, this will not affect the controllability of the vehicle. Driving inputs in normal driving are steering, throttle and braking. In order to reduce mental fatigue of human operator and eliminate coupling of decision parameters (steering and throttle), cruise speed control can be implemented inside the vehicle. Which inherently eliminates delay out of the control loop for longitudinal dynamics. This cruise control node receives reference speed from the human operator at the start of the teleoperation task or at regular time-interval. Consequently, the control input ($\delta$) is only the steering angle. In case of constant time-delays, the schematic given in fig (\ref{fig:smith_1}), can be simulated in continuous time domain, without maintaining any history of the states. But, to simulate variable downlink delays linked with output ($X$), as $(1-e^{-\left(\tau_{1}+\tau_{2}\right) s})$ suggests current output of $P^{\prime}$ has to be subtracted by its output at $(t - \left(\tau_{1}+\tau_{2}\right) )$. Time history of output of $P^{\prime}$ needs to be managed to be available when the delayed output is required.

For vehicle teleoperation simulation, $P^{\prime}$ is a dynamic single-track model, whose equations of motion are given by first seven terms of eq \ref{eq statesDot}, with null acceleration ($a=0$). P is the main vehicle with 14-dof vehicle model using MATLAB vehicle dynamics blockset \cite{vdynblks}.
\subsection{Successive reference-pose tracking using NMPC}

In vehicle teleoperation, model-based prediction approach as discussed in section above is effective, as dynamic single track vehicle model is a decent estimation in linear tire-dynamics range when lateral acceleration $<4m/s^2$ \cite{Reicherts2021} on normal roads and in normal conditions. But disturbances such as strong winds, very low adherence roads, bumps etc., can change the vehicle dynamics for small as well as for big duration. Parameter estimation techniques discussed in articles \cite{Teng, Thomas} are useful for medium to high duration change in dynamics. As this techniques are based on sliding window batch estimation, estimations has a lead time associated to it. Momentary disturbances can impact the vehicle output significantly before the new plant dynamics is estimated at the control station and corrective action taken by the human operator.

\begin{figure}[h]
\centering
    \includegraphics[width=\columnwidth]{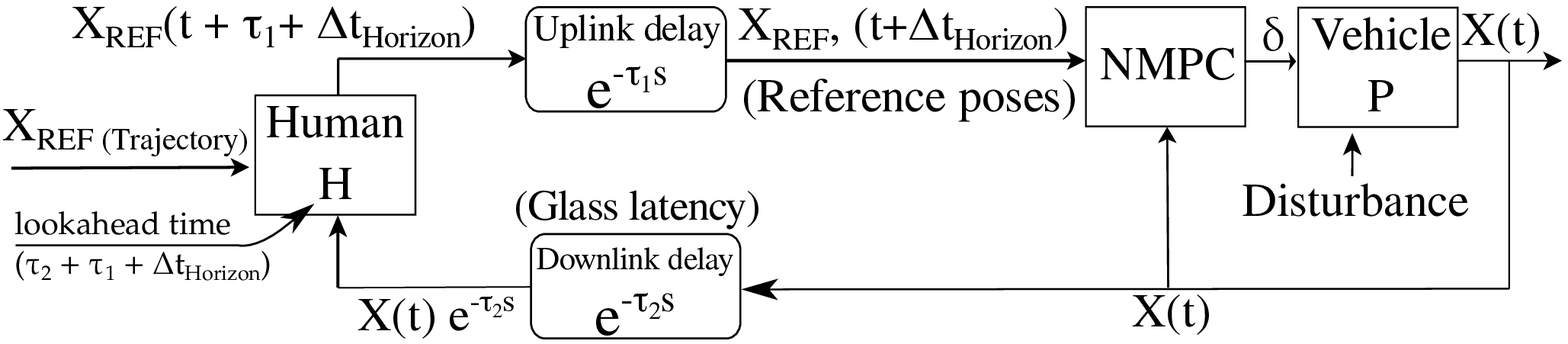}
    \caption{Successive reference-pose tracking (SRPT) scheme}
    \label{fig:mpcScheme}%
\end{figure}

In SRPT approach, instead of the control station transmitting the steering inputs, it transmits successive reference 2D-poses to be followed by the teleoperator vehicle. The schematic of SRPT approach in vehicle teleoperation control loop is shown in fig (\ref{fig:mpcScheme}).

The NMPC block receives the reference pose which belongs to the (approximate) end of its prediction horizon. Important fact is that it receives feedback (vehicle states, $X(t)$) without any network delay from the vehicle. By analyzing reference pose, vehicle current states, actuator and environmental constraints, it generates optimize steer and acceleration inputs. On the control-station side, the human operator receives delayed vehicle pose. Knowing the whole reference trajectory to be followed by the vehicle, downlink delay, uplink delay and prediction horizon, the operator transmits the reference pose for the end of the prediction horizon. Here the downlink delay is considered varying but known, uplink delay is considered a known constant and prediction time horizon is also a known constant.

\subsubsection{NMPC vehicle model}
\begin{figure}[h]
\centering
    \includegraphics[width=0.5\columnwidth]{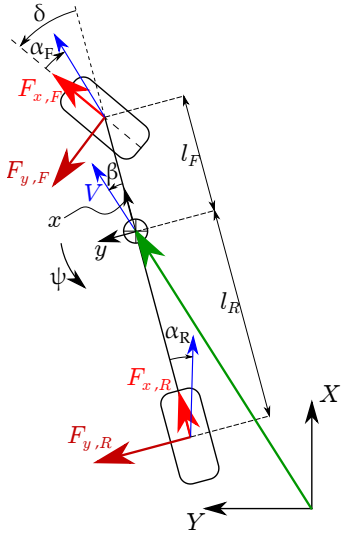}
    \caption{Single track vehicle model}
    \label{fig:07_singleTrack}%
\end{figure}

Dynamic single track vehicle model, being a good trade-off between simplicity for real-time implementation and accuracy for high-performance vehicle control, is used as prediction model in NMPC. It is illustrated in fig (\ref{fig:07_singleTrack}) with states

\begin{equation}
\mathbf{x}=\left[\:\beta ; \:\dot{\psi} ; \:\psi ; \:F_{y, F} ; \:F_{y, R} ; \:x ; \:y ; \:\delta ; \:V_x\right],\label{eq states}
\end{equation}

$\beta$ is side-slip angle, $[F_{y, F}, F_{y, R}]$ are lateral forces at axles in wheel reference plane, $V_x$ is vehicle longitudinal velocity, $\delta$ is steering angle. Before each optimization routine, starting vehicle 2D-pose resets $[x=0;\: y=0;\: \psi=0]$. This is because vehicle receives target reference pose in global reference frame, but before feeding it to NMPC, it is converted into vehicle reference frame.
\\

Balancing the lateral forces gives
\begin{equation}
m A_{x}=m\left(\dot{V}_{y}+\dot{\psi} V_{x}\right)=F_{y, F} \cos \delta+F_{x, F} \sin \delta+F_{y, R}\label{eq Fy1}
\end{equation}

Considering small $\beta$ angle

\begin{equation}
\begin{aligned}
&V_{x}=V \cos \beta \simeq V \\
&V_{y}=V \sin \beta \simeq V \beta
\end{aligned}\label{eq smallBeta}
\end{equation}

eq (\ref{eq Fy1}) becomes
\begin{equation}
\dot{\beta}=\frac{1}{m V}\left(F_{y, F} \cos \delta+F_{x, F} \sin \delta+F_{y, R}\right)-\frac{\beta \cdot a}{V}-\dot{\psi}\label{eq Fy2}
\end{equation}

Balancing the yaw moment around the center of mass gives
\begin{equation}
\ddot{\psi}=\frac{1}{I_Z}\left(\left(F_{y, F} \cos \delta+F_{x, F} \sin \delta\right)l_F - F_{y, R}l_R\right).\label{eq yaw1}
\end{equation}

For simplicity, relaxation length phenomenon is ignored in longitudinal force development and thus given by

\begin{equation}
\begin{aligned}
&F_{x, F}= 
\begin{cases}
    m\,a + f_V\,m_R\,g + C_{Aero}\,V^2,& \text{if } a\geq 0\\
    \:\:\:\:\:\:\:\:\:\gamma\,(m\,a + f_V\,m\,g + C_{Aero}\,V^2),              & \text{otherwise}
\end{cases} \\
&F_{x, R}= 
\begin{cases}
    \:\:\:\:\:\:\:\:\:-f_V\,m_R\,g,& \text{if } a\geq 0\\
    (1-\gamma)\,(m\,a + f_V\,m\,g + C_{Aero}\,V^2),              & \text{otherwise}
\end{cases}
\end{aligned}\label{eq Fxf}
\end{equation}

\begin{equation}
\begin{array}{l}
\zeta_F=\sqrt{1-\left(\frac{F_{x, F}}{\mu_{cons} \:m_{F}\:g}\right)^{2}}\\
\zeta_R=\sqrt{1-\left(\frac{F_{x, R}}{\mu_{cons} \:m_{R}\:g}\right)^{2}}
\end{array}
\end{equation}\label{eq zeta}

\begin{equation}
\begin{array}{l}
\alpha_{F}\simeq\tan ^{-1}\left(\tan \delta-\beta-\dot{\psi} \frac{l_F}{V}\right) \\
\alpha_{R}\simeq\tan ^{-1}\left(\:\:\:\:\:\:\:\:\:\:\:-\beta+\dot{\psi}\frac{ l_R}{V}\right)
\end{array}
\end{equation}\label{eq slips}

$[C_{\alpha, F}, C_{\alpha, R}]$ are the lumped cornering stiffness's of front and rear axles respectively; $[\zeta_F, \zeta_R]$ are the reduction factor for cornering stiffness considering longitudinally forces; $[\alpha_{F}, \alpha_{R}]$ are the slip angles of tires. [$m_f, m_R$] are distribution of vehicle mass on front and rear axle based of [$l_F, l_R$].

Altogether, the NMPC prediction model can be expressed as
\begin{equation}
\dot{\mathbf{x}}=\left[\begin{array}{c}
\frac{1}{m V}\left(F_{y, F} \cos \delta+F_{x, F} \sin \delta+F_{y, R}\right)-\frac{\beta \cdot a}{V}-\dot{\psi} \\
\frac{1}{I_Z}\left[\left(F_{y, F} \cos \delta+F_{x, F} \sin \delta\right)l_F - F_{y, R}\:l_R\right] \\
\dot{\psi} \\
\frac{V}{\lambda}\left[\zeta_F\,C_{\alpha, F} \:\alpha_{F}-F_{y, F}\right] \\
\frac{V}{\lambda}\left[\zeta_R\,C_{\alpha, R} \:\alpha_{R}-F_{y, R}\right] \\
V \cos (\psi+\beta) \\
V \sin (\psi+\beta) \\
\dot{\delta} \\
a
\end{array}\right].\label{eq statesDot}
\end{equation}

Here, $a$ is the input longitudinal acceleration, $\dot{\delta}$ is input steer velocity. To make it applicable for zero vehicle speed, wherever the $V$ is in denominator, is substituted by $max(0.01, V)$.

\begin{table}[h]
\centering
\caption{Vehicle parameters for the single-track model}
\begin{tabular}{|c|c|}
\hline
\textbf{Parameter} & \textbf{Value} \\ \hline
$m$                  & 1180\:kg         \\ \hline
$I_Z$                & 2066\:kg\,s$^2$    \\ \hline
$l_F$                & 1.515\:m         \\ \hline
$l_R$                & 1.504\:m         \\ \hline
$C_{\alpha, F}$      & 46\:kN/rad     \\ \hline
$C_{\alpha, R}$      & 46\:kN/rad     \\ \hline
$\lambda$ (Relaxation\,length)           & 0.3\:m              \\ \hline
$\gamma$ (Braking\,bias)            & 0.6              \\ \hline
$C_{Aero}$ (Aerodynamic\,drag)            & 0.4\:N\:/\:(m$^2$/s$^2$)              \\ \hline
$f_V$ (Rolling\,resistance\,coeff)            & 0.025              \\ \hline
\end{tabular}
\label{tab:vehicleParameters}
\end{table}

\subsubsection{NMPC constraints}
Input constraints are steering velocity, vehicle acceleration and deceleration.
\begin{equation}
\begin{array}{c}
\left[\dot{\delta}_{min}=-10^{\circ}/s\right] \leq \dot{\delta} \leq \left[\dot{\delta}_{max}=+10^{\circ}/s\right] \\
\:\:\:\:\left[a_{min}=-\mu_{cons}\cdot4m/s^2\right] \leq a \leq \left[a_{max}=\mu_{cons}\cdot1m/s^2\right]

\end{array}
\end{equation}\label{eq inputConstraints}
$[a_{min}, a_{max}]$ are down-scaled by $\mu_{cons}$, to prohibit friction force saturation on low adhesion surface. Output constraints are max steering angle, vehicle speed, friction utilization at the tires. They can be summarized as 

\begin{equation}
\begin{array}{c}
\left[\delta_{min}=-25^{\circ}\right] \leq \delta \leq \left[\delta_{max}=+25^{\circ}\right] \\
0 \leq V \\
\frac{\left\|\left(\zeta_{F} C_{\alpha, F} \,\alpha_{F}, \:\:F_{x, F}\right)\right\|_{2}}{m_{F} \,g} \leq \mu_{cons} \\
\frac{\left\|\left(\zeta_{R} C_{\alpha, R} \,\alpha_{R}, \:\:F_{x, R}\right)\right\|_{2}}{m_{R} \,g} \leq \mu_{cons}.
\end{array}
\end{equation}\label{eq constraints}

The fact that human is in the loop, can be exploited in assigning a conservative friction coefficient $\mu_{cons}$. E.g., if the upcoming road is dry asphalt or snowy, the human operator assigns a conservative value of $\mu_{cons}=0.90$ or $\mu_{cons}=0.25$ to NMPC respectively.

\subsubsection{NMPC cost function}
The primary aim is to align the target reference pose with optimized vehicle trajectory. Clothoid path is a preferred path in vehicle motion, because it resembles natural driving where steer changes linearly. Eliou and Kaliabetsos \cite{Eliou2014ANS}, suggests cubic spline can be a first approximation for clothoid curve. For each prediction NMPC node estimates a cubic spline (eq \ref{eq spline1}-\ref{eq spline2}) from vehicle CG to reference pose in vehicle reference frame.

\begin{equation}
y=A x^{3}+B x^{2}+C x+D \label{eq spline1}
\end{equation}

\begin{figure}[h]
\centering
    \includegraphics[width=0.5\columnwidth]{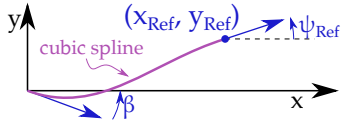}
    \caption{Cubic spline generation; inputs are in blue.}
    \label{fig:08_spline}%
\end{figure}

Target reference pose in vehicle reference frame is $[x_{Ref}; \,y_{Ref}; \,\psi_{Ref}]$, the cubic coefficients are given by
\begin{equation}
{\left[\begin{array}{l}
A \\
B \\
C \\
D
\end{array}\right]=\left[\begin{array}{cccc}
0 & 0 & 0 & 1 \\
x_{Ref}^{3} & x_{Ref}^{2} & x_{Ref} & 1 \\
0 & 0 & 1 & 0 \\
3 x_{Ref}^{2} & 2 x_{Ref} & 1 & 0
\end{array}\right]^{-1}\left[\begin{array}{c}
0 \\
y_{Ref} \\
\tan \beta \\
\tan \psi_{Ref}
\end{array}\right]}\label{eq spline2}
\end{equation}

which maintains $G1$ continuity at both ends, i.e., starting tail with $\beta$ and end tail of spline with $\psi_{Ref}$, as shown in fig \ref{fig:08_spline}.

Cost function of NMPC formulation is given by

\begin{equation}
\begin{aligned}
\min \quad &  \sum_{i=0}^{N-1}{U_i^{t}\,R\,U_i} + \sum_{i=0}^{N-1}{X_i^{t}\,Q\,X_i} + X_N^{t}\,P\,X_N\\
\end{aligned}\label{eq cost0}
\end{equation}

Where,
\begin{flalign}
U_{i}&=\left[\begin{array}{l}
\dot{\delta} \\
a
\end{array}\right] \quad \forall i \in[0, N-1]&\label{eq13}
\\
X_{i}&=\left\{\begin{array}{l}
{\left[V_{Ref}-V_{i}\right] \quad \forall i \in[0, N-1]} \\\\
{\left[\begin{array}{l}
A x_i^{3}+B x_i^{2}+C x_i+D\:\:\:\:\:\:-\:y_i \\
\tan ^{-1}\left(3 A x_i^{2}+2 B x_i+C\right)-\psi_i
\end{array}\right] \quad \forall i=N}
\end{array}\right.&\label{eq15}
\end{flalign}

The stage cost ($\forall i \in[0, N-1]$) tries to minimize inputs ($U_i$) and also tries to be as near as possible to the reference vehicle speed ($V_{Ref}$) asked by the human operator. The first term in terminal cost aims to coincide end point of the NMPC trajectory with the reference spline and second term match the heading of end point with spline tangent. $[x_i, y_i]$ are trajectory points in vehicle reference frame. Penalties assigned in the cost function is summarized in the Appendix. Prediction horizon ($\Delta t_{Horizon}$) is of $1\,sec$. The optimal control problem is discretized in $N=50$ intervals using discrete multiple shooting and solved by sequential quadratic programming using the real-time NMPC solver ACADOS \cite{Verschueren2019, Verschueren2018}.
\section{Simulation Results}

\begin{figure}[h]
\centering
    \includegraphics[width=0.8\columnwidth]{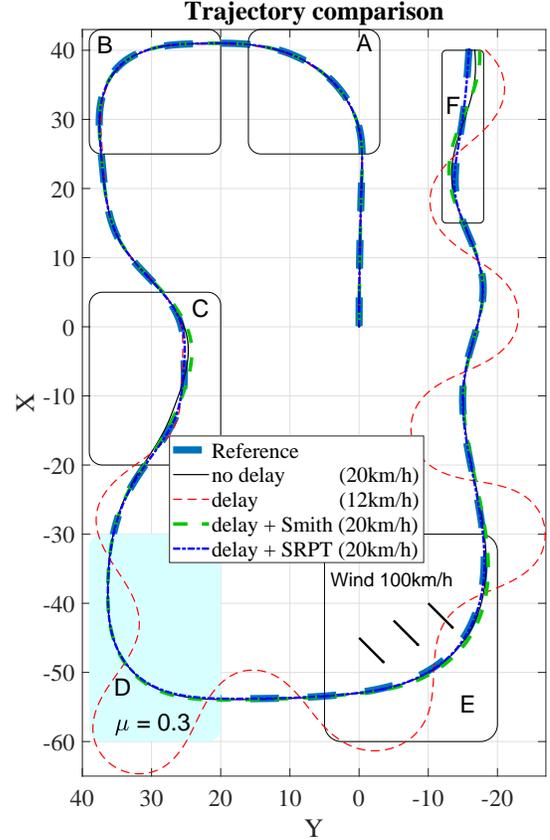}
    \caption{Trajectory tracking comparison between smith predictor approach and SRPT approach at V$_{Ref}$=20km/h}
    \label{fig:08_trajectoryComparison}%
\end{figure}

To compare the smith predictor approach and SRPT approach, \textit{Reference} trajectory shown in figure \ref{fig:08_trajectoryComparison} is chosen. It has 5 sections, namely A-F. A-B is cornering, C is double lane change, D is cornering on low adhesion road, E is cornering in strong diagonal wind and F is slalom. For simulation, vehicle dynamics blockset \cite{vdynblks} of MATLAB is utilized. It consists of a 14-dof passenger FWD vehicle model which accepts steering, acceleration and brake commands. Longitudinal dynamics control is kept same for the comparison, which is a PI control with tracking windup and feed-forward gains to maintain cruise speed of $20km/h$. Artificial bidirectional delay is introduced to simulate teleoperation network delays, constant uplink-delay ($\tau_1$) of $60ms$ and variable downlink-delay ($\tau_2$) as discussed in section \rom{2-A}.

Human ($H$) in smith predictor approach (fig \ref{fig:smith_1}) is simulated by kinematic Stanley controller \cite{stanley2007}.
\begin{equation}
\delta(t)= \begin{cases}\psi(t)+\tan^{-1} \frac{k e(t)}{V(t)} & \text { if }\left|\psi(t)+\tan^{-1} \frac{k e(t)}{V(t)}\right|<\delta_{\max } \\ \:\:\:\delta_{\max } & \text { if } \psi(t)+\tan^{-1} \frac{k e(t)}{V(t)} \geq \delta_{\max } \\ -\delta_{\max } & \text { if } \psi(t)+\tan^{-1} \frac{k e(t)}{V(t)} \leq-\delta_{\max }\end{cases} \label{eq:stanley}
\end{equation}

Here, $\psi(t)$ is the yaw angle (heading) of the vehicle with respect to the closest trajectory segment. $e(t)$ is cross-track error from the front axle center. $V(t)$ is the vehicle speed in $m/s$ and $k$ is tuneable gain which is set to be equal to $0.7$ in this case.

Whereas the human ($H$) in SRPT approach (fig \ref{fig:mpcScheme}) is a simple look-ahead pose selector, which receives delayed pose and vehicle speed. Knowing the target reference trajectory, vehicle delayed pose and speed, it selects and transmits the look ahead pose corresponds to [$\tau_2+\tau_1+\Delta t_{Horizon}$] seconds ahead of the delayed pose. The NMPC controller runs well below 20 ms ($mean = 6ms, max = 15ms$) on an Intel Core i7-11800H 4.6 Ghz 8 Core CPU with 16 GB of RAM in parallel with Simulink environment and hence it can be expected to perform real-time @50Hz on a dedicated hardware of the experimental vehicle.

\begin{figure}[h]
\centering
    \includegraphics[width=1.0\columnwidth]{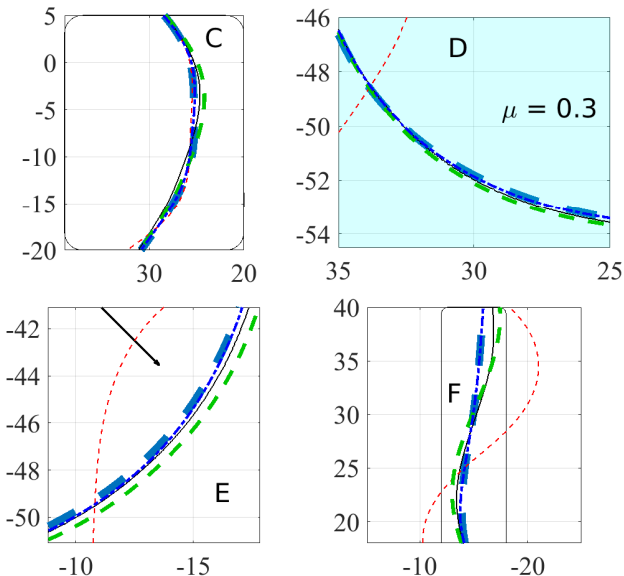}
    \caption{Zoomed regions}
    \label{fig:08_zoomedRegions}%
\end{figure}

Trajectories traversed by the two approaches smith predictor and SRPT is presented in figure \ref{fig:08_trajectoryComparison}. For better comparison, trajectory traversed in case of no delay and delay with no teleoperation control is also presented. Due to poor path following performance in case of delay with no teleoperation control, only for this particular mode, low vehicle speed of $12 km/h$ is chosen. The zoom plots in fig \ref{fig:08_zoomedRegions}, emphasise better reference tracking behaviour with the SRPT approach in all sections of the trajectory. In this figure, region A and B are ignored, due to insignificant improvement found. 

\begin{figure}[h]
\centering
    \includegraphics[width=1.0\columnwidth]{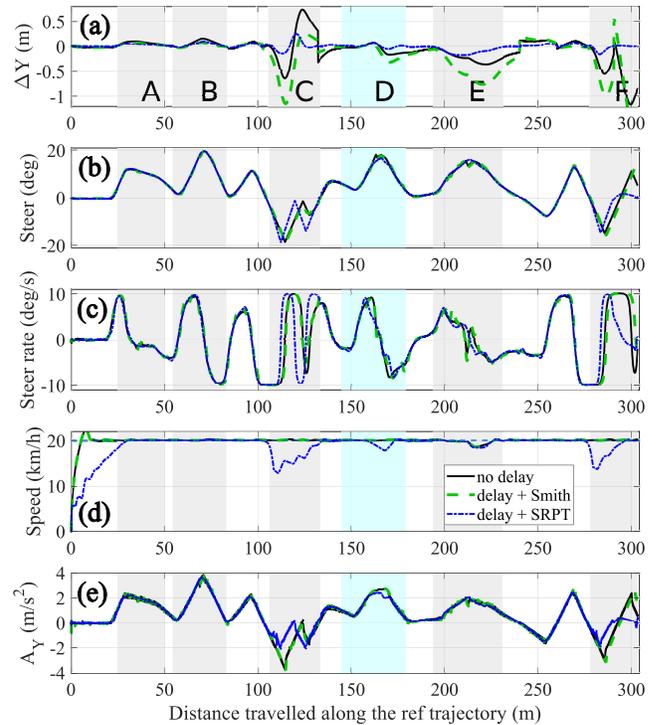}
    \caption{Cross-track error, steer, steer-rate, vehicle speed, and lateral acceleration comparison}
    \label{fig:09_performanceComparison}%
\end{figure}

Figure \ref{fig:09_performanceComparison}a quantifies the performance difference by comparing the RMS of lateral deflection from the reference trajectory for each section of the trajectory.

RMS deflection is computed as given by
\begin{align}
\Delta Y_{\mathrm{RMS}}=\sqrt{\frac{1}{D_{i+1}-D_i} \int_{D_i}^{D_{i+1}}[\Delta Y(D)]^{2} \: dD}\label{eq34}
\end{align}   

where,
\\
$\Delta Y(D)$ - is the lateral deflection of vehicle CG from the reference trajectory
\\
$D$ - distance travelled along the reference trajectory.
\\
$i, i+1$ - indicates the start and end of the particular region (fig \ref{fig:09_performanceComparison}a).

Compare to the deflections observed with smith predictor approach, deflections observed with SRPT are lesser in all the trajectory sections except for region B, where difference is insignificant as shown in fig \ref{fig:09_performanceComparison}a and fig \ref{fig:10_bars}. Figure \ref{fig:09_performanceComparison}(b-c-d) shows steer, steer rate, vehicle speed, and lateral acceleration respectively vs the distance travelled. 

Result is analysed below for each trajectory section:
\subsection*{Trajectory section - A, and B}
Insignificant differences are found in all the teleoperation modes due to less aggressive steer requirement due to low curvature.

\subsection*{Trajectory section - C, and F}
Successive pose tracking approach showed lesser deviations because wherever the steer-rate requirement exceeds actuator limitation ($\:|\dot{\delta}| \leq \;\; 10 deg/s\:$), NMPC moderates the vehicle speed (fig \ref{fig:09_performanceComparison}d) to allow more time for steering. It can be clearly observed in C and F trajectory sections, where steer (fig \ref{fig:09_performanceComparison}b) is actuated slightly before in distance as compare to smith predictor approach.

\subsection*{Trajectory section - D}
Here the road persists low adhesion property ($\mu = 0.3$), which corresponds to snowy environment condition. Exploiting the fact that human is in the control loop, human can convey a conservative friction ($\mu=0.25$ in this case) value to the NMPC controller at the section entrance. According to which, NMPC moderated the vehicle speed and started steering in advance (fig \ref{fig:09_performanceComparison}c) to be on the reference track. On the contrary smith predictor estimation accuracy suffered due to the presence of disturbance in the plant and eventually resulted noisy steering command.

\subsection*{Trajectory section - E}
Here in the presence of strong cross winds ($V_{Wind} = 100 \;km/h$), again disturbance in the plant causes noisy steering commands (fig \ref{fig:09_performanceComparison}c) with smith predictor approach. Whereas being unaware of cross wind disturbance but aware of reference poses, NMPC steering commands are smooth. Which leads to less deviations.

\begin{figure}[h]
\centering
    \includegraphics[width=1.0\columnwidth]{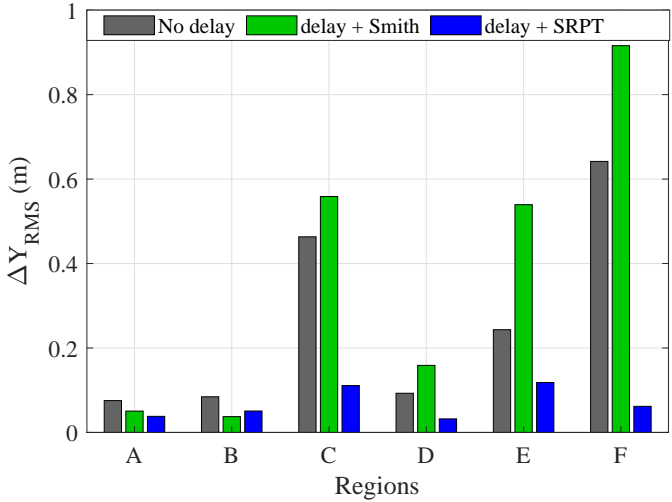}
    \caption{Cross-track error comparison in individual regions with different teleoperation modes}
    \label{fig:10_bars}%
\end{figure}

Figure \ref{fig:10_bars}, summarizes the cross-track error in all regions with all modes. Shortness of blue bars with SRPT mode, reflects significant less lateral error in vehicle teleoperation.
\section{Conclusion}
The classical Smith predictor approach to mitigate detrimental time-delay effects in closed-loop operation is compared with proposed SRPT approach for vehicle teleoperation in presence of disturbances such as low adhesion road, cross-wind and aggressive maneuvers such as slalom \& double-lane change. The simulation framework consists of Simulink environment with 14-dof vehicle model equipped with PI cruise control feature. Smith predictor approach transmits directly the steer commands  to the vehicle, whereas SRPT approach transmits reference poses to the vehicle based on look-ahead time and thereupon NMPC controller modulates steer as well as reference vehicle speed. The optimal routine uses ACADOS frameowrk, which shows the capability to run at $50Hz$ as mean computation time observed is $6ms$.

The optimization routine penalizes cross-track error and eventually decelerate the vehicle to dilate the time-window available for steering to account saturation of steering actuator. To analyse the performance benefit, RMS of cross-track error is compared for different critical trajectory sections shown in figure \ref{fig:08_trajectoryComparison}. Comparison indicates significant improvement in terms of reduction in cross-track error (fig \ref{fig:10_bars}) with SRPT approach.

\textit{Futurework} - This paper presents a control approach where successive reference poses are transmitted to the vehicle at close time intervals (@30fps) as compare to transmitting steering commands directly to the vehicle. Successive work would propose a novel method where the issue of real-time generation of successive reference poses is attempted to solve with joystick steering wheel to provide a realistic driving experience to the human operator. A human in the loop vehicle simulator environment would be utilized to assess performance benefits with the proposed SRPT approach.
\appendices
\section*{Appendix}
NMPC cost penalties are summarized in the following table.

\begin{table}[h]
\centering
\caption{NMPC cost penalties}
\begin{tabular}{|c|c|l|}
\hline
\textbf{R}     & \textbf{Q} & \multicolumn{1}{c|}{\textbf{P}} \\ \hline
$diag([1, 0.1])$ & $1$          & $diag([50, 3])$                   \\ \hline
\end{tabular}
\label{tab:mpcWeights}
\end{table}

\ifCLASSOPTIONcaptionsoff
  \newpage
\fi

\bibliographystyle{IEEEtran}
\bibliography{sample}



\end{document}